\pdfoutput=1

\documentclass[11pt]{article}

\usepackage[preprint]{acl}

\usepackage{times}
\usepackage{latexsym}

\usepackage[T1]{fontenc}

\usepackage[utf8]{inputenc}

\usepackage{microtype}

\usepackage{inconsolata}

\usepackage{graphicx}

\usepackage{subcaption}
\usepackage{algorithm}
\usepackage{algorithmic}
\usepackage{amsmath}
\usepackage{svg}
\usepackage{amsthm}   
\usepackage{amssymb}
\usepackage{listings}

\title{Inverse-Q*: Token Level Reinforcement Learning for Aligning Large Language Models without Preference Data}

\author{
    Han Xia$^{1}$\thanks{{ }Equal contribution}, \ \ Songyang Gao$^{2 \star}$, \ \ Qiming Ge$^{1}$, \ \ Zhiheng Xi$^{1}$, \ \ Qi Zhang$^{1}$, \ \ Xuanjing Huang$^{1}$\thanks{{ }{}{}Corresponding author},\\
    \normalsize{$^1$ School of Computer Science, Fudan University, Shanghai, China} \\
    \normalsize{$^2$ Shanghai Artificial Intelligence Laboratory, Shanghai, China} \\
    \normalsize{hxia22@m.fudan.edu.cn, gaosongyang@pjlab.org.cn} \\
}

\begin{document}
\maketitle
\begin{abstract}
Reinforcement Learning from Human Feedback (RLHF) has proven effective in aligning large language models with human intentions, yet it often relies on complex methodologies like Proximal Policy Optimization (PPO) that require extensive hyper-parameter tuning and present challenges in sample efficiency and stability. In this paper, we introduce Inverse-Q*, an innovative framework that transcends traditional RL methods by optimizing token-level reinforcement learning without the need for additional reward or value models. Inverse-Q* leverages direct preference optimization techniques but extends them by estimating the conditionally optimal policy directly from the model's responses, facilitating more granular and flexible policy shaping. Our approach reduces reliance on human annotation and external supervision, making it especially suitable for low-resource settings. We present extensive experimental results demonstrating that Inverse-Q* not only matches but potentially exceeds the effectiveness of PPO in terms of convergence speed and the alignment of model responses with human preferences. Our findings suggest that Inverse-Q* offers a practical and robust alternative to conventional RLHF approaches, paving the way for more efficient and adaptable model training approaches.
\end{abstract}

\section{Introduction}

Reinforcement Learning from Human Feedback (RLHF, \citealp{Christiano_Leike_Brown_Martic_Legg_Amodei_2017}) is a mainstream approach for aligning large models to human intentions, demonstrated in applications such as ChatGPT \citep{ouyang2022training} and Llama3 \citep{llama3}. The RLHF framework involves modeling a reward function from preference data and learning an optimal policy through PPO \citep{schulman2017proximal}, which also estimates expected returns, translating language modeling into an MDP problem. This method provides nuanced supervision over training samples, proving effective in tasks like instruction following and safety \citep{ramamurthy2022reinforcement, ouyang2022training, glaese2022improving}. Nonetheless, PPO's high performance depends on complex optimization and parameter tuning, raising concerns about its sample efficiency and stability.

\begin{figure}[t]
  \includegraphics[width=\columnwidth]{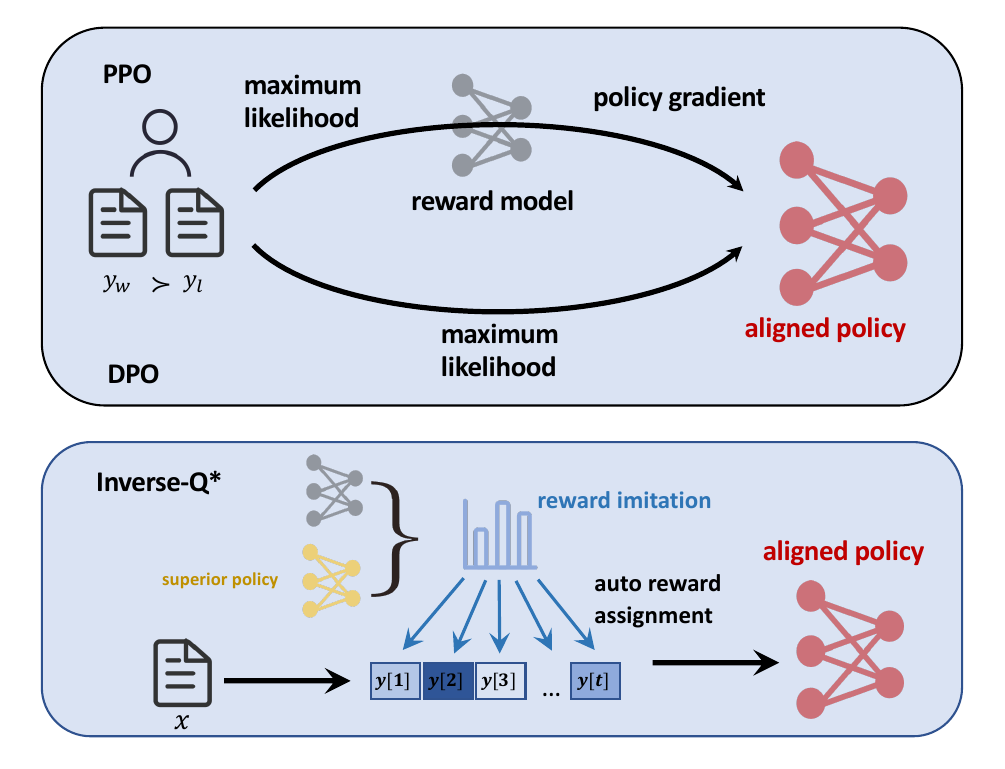}
  \caption{Existing model alignment approaches require preference data for reward modeling. However, Inverse Q* utilize reward imitation from superior strategies to achieve token-level credit assignment, making model alignment more efficient without preference data.}
  \label{fig:mainfigure}
\end{figure}

As an efficient alternative to PPO, Direct Preference Optimization (DPO, \citealp{rafailov2024direct}) aligns large models from the perspective of contextual bandits, not token-level decisions (\citealp{yue2012k}, \citealp{dudik2015contextual}). DPO optimizes preference reward loss directly through reward model loss, affecting the probability margins of preference pairs. Similar methods like RSO \citep{tripathi2020rso}, ReST \citep{gulcehre2023reinforced}, and ReST-em \citep{singh2023beyond} train policies to fit optimal prior distributions on predefined response sets, avoiding the need for a critic model. However, these methods still require additional supervisory signals, such as a reward model, to enhance response quality, leading to trade-offs in labeling costs and accuracy. While direct optimization methods generally overlook token-level preference modeling, some efforts (\citealp{chan2024dense}) have explored using reward assignment to refine feedback signals, though these enhancements mainly bolster method stability rather than guide updates.

A crucial observation is that direct optimization methods still require the logits of entire response sequences to construct the loss function due to the need for differentiability in back-propagation. Lacking corresponding advantage function modeling, such constructions cannot naturally generalize to token-level process supervision. Based on this insight, we hypothesize: \textit{Is there a special trajectory estimation whose feedback signal can naturally generalize to dense reward function modeling within the token MDP, thereby automatically constructing advantage function interpretations for each token?}

Similar properties were demonstrated in r2Q*\citep{rafailov2024r}, where DPO training could implicitly learn the optimal Q-function and mimic controllable decoding, but it required pre-labeled preference data for obtaining the reference distribution under the optimal policy. In this work, we estimate the conditionally optimal distribution for current inputs on single dialogue data without additional labeling or external supervision. We introduce \textbf{Inverse-Q*}, an algorithm that optimizes the same objective as PPO (maximizing the advantage function) with enhanced flexibility and easier implementation. Our method, an inverse problem of DPO training, assigns token-level reward feedback via an estimated policy, optimizing the large model online within the MDP framework.

Overall, Inverse-Q* exhibits similar sample utilization efficiency and supervision granularity as PPO, providing token-level RL training across all sampling outcomes without relying on additional reward models or value models, thus performing excellently in terms of labeling and computational resource demands. The process of Inverse-Q* is illustrated in Figure 1, and we have conducted extensive experiments to demonstrate the efficacy of our framework in low-resource RLHF training. Inverse-Q* has shown the capability to achieve or even exceed the effectiveness of PPO training. Our contributions can be summarized as follows:

\begin{enumerate}
    \item We introduce Inverse-Q*, a novel framework that estimates the optimal policy under current problems, offering improved convenience and flexibility.
    \item We demonstrate the reliability of our framework through rigorous proofs, and provide a corresponding practical algorithm based on Inverse-Q*, which performs token-level reinforcement learning without preference labeling or external supervision.
    \item Empirical studies show that our method significantly improves the alignment of large language model responses with human preferences compared to other RLHF methods, and achieves faster convergence relative to PPO and DPO training.
\end{enumerate}

\section{Related Works}

\subsection{Reinforcement Learning from Human Feedback}

Aligning policy models with objectives is crucial in reinforcement learning. RLHF algorithms, particularly those using the PPO algorithm with a KL penalty, are mainstream for aligning language models. These methods optimize a reward model on preference data and employ on-policy reinforcement with PPO, which also trains a critic model (value model) to estimate future rewards. This approach has improved response accuracy, reduced harmful content, and adjusted response styles but faces challenges like optimization instability and high computational demands \citep{Christiano_Leike_Brown_Martic_Legg_Amodei_2017, ouyang2022training}.

\subsection{Credit Assignment}

Exploration with sparse rewards is challenging. Credit Assignment methods distribute supervisory signals sentence-wise and optimize with PPO, enhancing training stability and learning speed. Attention Based Credit (ABC, \citep{chan2024dense}) redistributes rewards token-wise using attention weights from the reward model. Reinforced Token Optimization (RTO, \citep{zhong2024dpo}) and r2Q* \citep{rafailov2024r} derive DPO at the token-level MDP, demonstrating effective credit assignment.

\subsection{Self-Improvement}

Obtaining high-quality human data is resource-intensive. RL from AI Feedback (RLAIF, \citep{bai2022constitutional}) uses model-generated synthetic data, drastically cutting costs by requiring minimal human supervision. Reinforced Self-Training (ReST, \citep{gulcehre2023reinforced}) and ReST-EM \citep{singh2023beyond} iterate on policy-sampled data, refined by a reward function, for enhanced model training. Our method can also be viewed as a self-improvement approach, but it neither relies on external feedback nor requires an additional trained reward model to delineate the optimal strategy. Instead, it optimizes based on the model’s own estimation of the optimal strategy.

\section{Preliminaries}

In this section, we first introduce the classical RLHF framework in LLM alignment, followed by a description of how this modeling is tied to direct alignment methods (in the case of DPO). Given a prompt $\mathbf{x}^*$ sampled from the dataset $\mathcal{D}=\left\{\left(\mathbf{x}_{i}, \mathbf{y}_{i}\right)\right\}_{i=0}^N$, policy model provides a multi-token response \( \mathbf{y}^* = (y_0, \ldots, y_T) \) to complete a full interactive dialogue process. To align with the output format of language models throughout this chapter, we use \( \mathbf{y}_{t-1} = (\mathbf{x}^*, y_0, \ldots, y_{t-1}) \) to denote the current state at time \( t \) in the RL context, where \( y_t \) represents the policy action at token level.

Most RLHF algorithms require training a reward function from human preference data to provide online feedback on model outputs. A preference data pair \( (x, y_w, y_l) \) typically begins with the same initial prompt and receives a corresponding reward score at termination, and the probability of preferring \( \tau^w \) over \( \tau^l \) is given by:

{\small
\begin{align}
p^*\left(\mathbf{y}^w \succeq \mathbf{y}^l\right) = \frac{\exp \left( r\left(\mathbf{x}^w, \mathbf{y}^w\right)\right)}{\exp \left( r\left(\mathbf{x}^w, \mathbf{y}^w\right)\right) + \exp \left( r\left(\mathbf{x}^l, \mathbf{y}^l\right)\right)},
\end{align}}

where \( r(\mathbf{x}, \mathbf{y}) \) denotes the reward function for state-action pair. 

This modeling is subsequently used to optimize the generate policy of LLMs by improving the preferring probability of model responses over older ones. However, human preference annotations typically only exist at the response or sentence level, so the reward model cannot directly provide gradient signals action by action for optimization. \textbf{PPO} artificially defines token-level rewards with entropy bonuses to adhere to the Bradley-Terry preference modeling \citep{bradley1952rank} as follows:

{\small\begin{align}\label{reward}
    r\left( \mathbf{y}_t\right)= \begin{cases}\beta \log \pi_{\mathrm{ref}}\left(\mathbf{y}_t \mid \mathbf{y}_{t-1}\right), & \text { if not end} \\ r(\mathbf{y_t})+\beta \log \pi_{\mathrm{ref}}\left(\mathbf{y}_t \mid \mathbf{y}_{t-1}\right), & \text { if end } \end{cases}
\end{align}}

Based on the above definition, PPO aims to maximize the expected reward at each token while ensuring that the learned policy does not diverge significantly from a reference model. For a given input \( \mathbf{y}_{s} \), the optimal policy is represented as:

{
\small
\begin{align}\label{mdp}
\arg\max_{\pi} \mathbb{E}_\pi\left[\left.\sum_{t=s}^{T}\left(r\left(\mathbf{y}_t\right)-\beta \cdot \log \frac{\pi\left(\mathbf{y}_t \mid \mathbf{y}_{t-1}\right)}{\pi_{\mathrm{ref}}\left(\mathbf{y}_t \mid \mathbf{y}_{t-1}\right)}\right) \right\rvert\, \mathbf{y}_s\right]
\end{align}
}

where \( \beta \) is a parameter that balances reward and entropy bonuses, and \( \pi\left(\mathbf{y}_t \mid \mathbf{y}_{t-1}\right) \) is the policy's probability of choosing tokens.

On the contrary, DPO utilizes a contextual bandits setting to circumvent token-level reward allocation issues. Assuming an implicit reward model \( r \) that scores all potential responses \( \{ \mathbf{y}^*_i \}_{i=1}^m \) under prompt \( x^* \), the closed-form solution of the policy model under a KL-constrained contextual bandit optimization problem can be expressed as:

\begin{align}\label{reward score}
\pi^*(\mathbf{y} \mid \mathbf{x}) = \frac{1}{Z(\mathbf{x})} \pi_{\mathrm{ref}}(\mathbf{y} \mid \mathbf{x}) \exp\left( r(\mathbf{x}, \mathbf{y}) \right), 
\end{align}

where \( Z(\mathbf{x}) \) is a partition function. Reversing this conclusion, we obtain the reward modeling in current policy optimization (DPO) as:
\[ r(\mathbf{x}, \mathbf{y}) = \beta \log \frac{\pi^*(\mathbf{y} \mid \mathbf{x})}{\pi_{\mathrm{ref}}(\mathbf{y} \mid \mathbf{x})} - Z(\mathbf{x}), \]
This modeling is subsequently used to compute standard reward model losses for updating policy distributions.

\section{Methods}

We have analyzed the reward modeling of model responses in PPO and DPO in the previous section. In this section, we aim to develop a novel strategy optimization method that can provide fine-grained supervision for token-wise MDP problems without relying on external feedback. 

Our derivation starts from the optimization objective of PPO in Eq. \ref{mdp}, which can be viewed as Monte Carlo sampling from any state \( y_s \), aimed at estimating the value function under given state. We first demonstrate that fitting a superior policy on the reply space with complete reward function annotations can enhance expected returns, thereby inducing better alignment. Subsequently, we introduce the process of generalizing this approach from complete responses to the token level.

\subsection{Policy Optimization Through Reward Imitation}

For clarity, let's isolate the part related to the current policy from equation Eq. \ref{mdp}, and the alignment objective under the KL constraint can be expressed as: 

{\small\begin{align}
V(\pi; \mathbf{y}_s) = \mathbb{E}_{y \sim \pi\left(\cdot \mid \mathbf{y}_s\right)}\left[\sum_{t=s}^T r\left(\mathbf{y}_t\right) + \beta \mathcal{H}\left(\pi_\theta\right) \mid \mathbf{y}^s \right],
\end{align}}

where the definition of \( r \) follows Eq. \ref{reward}, and \( \mathcal{H}(\pi) \) denotes the entropy of the distribution \( \pi \). We propose the following lemma:
\newtheorem{theorem}{Theorem}[section]
\newtheorem{definition}{Definition}[section]
\newtheorem{lemma}{Lemma}[section]
\begin{lemma}[Reward Imitatioin]\label{lemma:1}
Considering two policies \( \pi_a \) and \( \pi_b \) where \( \pi_a \) is superior, meaning \( V(\pi_a; \mathbf{y}_s) > V(\pi_b; \mathbf{y}_s) \), for any imitation policy \( \pi_\theta = (1-\delta) \cdot \pi_b + \delta \cdot (\pi_b - \pi_a) \), where \( \delta \) is any real number in the interval (0, 1), it holds that \( V(\pi_\theta; \mathbf{y}_s) > V(\pi_b; \mathbf{y}_s) \).
\end{lemma}

\begin{proof}\let\qed\relax
Clearly, \( \pi_\theta \) is a probability distribution over the same state space as \( \pi_a \) and \( \pi_b \).
Continuing from Eq. 5, since the entropy of policy \( \pi \) is independent of the actual sampling of generated results, we have:

{\small\begin{align*}
    & \ \ \ \  V(\pi_a; \mathbf{y}_s) - V(\pi_b; \mathbf{y}_s) \\
    &= \beta (\mathcal{H}(\pi_a) - \mathcal{H}(\pi_b)  ) +\int_{\mathcal{Y}} (\pi_a(y) -\pi_b(y)) r(\mathbf{y})dy  \\
    &\leq \frac{1}{\delta}\left( \beta (\mathcal{H}(\pi_\theta) - \mathcal{H}(\pi_b)  ) + \int_{\mathcal{Y}} (\pi_\theta(\mathbf{y}) -\pi_b(\mathbf{y})) r(\mathbf{y})d\mathbf{y} \right) \\
    &= \frac{1}{\delta}(V(\pi_\theta; \mathbf{y}_s) - V(\pi_b; \mathbf{y}_s)),
\end{align*}}
\end{proof}

The second transformation utilizes the concavity of the entropy function. Lemma \ref{lemma:1} states that when training towards a distribution direction given by a superior strategy, the model always yields better outcomes on in-domain data. 

This optimization process is similar to DPO with similar reward modeling provided under the comparison between the policy model and the reference model, which is shown in Eq. \ref{reward score}.
The distinction is that while DPO attempts to maximize the differences in generation probabilities between preference data to optimize the policy model, Reward Imitation uses a pre-estimated superior distribution to allocate confidence to given responses, subsequently adjusting the current policy's distribution to align with it. We naturally hope that Reward Imitation can automatically generalize to decision-making processes on individual tokens, thereby allowing us to directly use supervised fine-tuning to optimize the policy model (using the estimated token probabilities as soft labels). The optimized loss function would then be the value function corresponding to each token. However, as described in Eq. \ref{reward}, the reward feedback in large model alignment tasks is delayed, thus requiring extensive sampling, or an additional critic model to obtain value estimates for process tokens.

To address this issue, the optimal strategy is to select a class of reward function whose output process precisely equals the expected future return, i.e., $r(y_t) = \mathbb{E}_{y \sim \pi(\cdot \mid \mathbf{y}_t)} [\pi(\mathbf{y})r(\mathbf{y})] $. In the next section, we will explain how our reward modeling naturally satisfies the above requirements, thus enabling RL training on individual tokens without value models.

\subsection{Reward Imitation Performs Auto Reward Assignment}

In the previous section, we presented our optimization algorithm called Reward Imitation, which estimates a superior strategy to allocate generation probabilities for current trajectories, thus aligning preferences on non-preference data. When extending this process to any intermediate step rather than just the termination state (i.e., EOS token), consistency between the reward function and the Q-function must be maintained. We now demonstrate that, when using a specific form of reward modeling, our estimated trajectory generation probabilities can naturally extend to any of their prefix sequences. \citet{rafailov2024r} and \citet{chan2024dense} have discussed the automatic construction of implicit Q-functions when preforming DPO training with paired preference data. Our operation can be viewed as the reverse of their process, which uses a temporarily estimated superior strategy on the current prompt to directly provide value scores for specific prefixes.

Given an arbitrary reply prefix \( y_{t-1} \) and an estimated superior strategy \( \pi^*(\cdot \mid \mathbf{x}) \) for that state, we define
\begin{align} \label{eq.estimate}
V(\pi^*(\cdot \mid \mathbf{x}), \mathbf{y}_t) = \beta \sum_{i=1}^t \log \frac{\pi^*\left(\mathbf{y}_i \mid \mathbf{y}_{<i}\right)}{\pi_{\mathrm{ref}}\left(\mathbf{y}_i \mid \mathbf{y}_{<i}\right)},
\end{align}
where \( \beta \) is the weight of the KL constraint, and \( \pi_{\mathrm{ref}} \) serves as the baseline model to provide a measure of the extent of policy changes. 

since both \( \pi^* \) and \( \pi_{\mathrm{ref}} \) are probability distributions over any response sequence and its prefixes, when sampling from the distribution \( \pi_{\mathrm{ref}} \) to estimate the value function using Monte Carlo methods, we have:

{\small\begin{align}\label{r2q}
&\beta \sum_{i=1}^t \log \frac{\pi^*\left(\mathbf{y}_i \mid \mathbf{y}_{<i}\right)}{\pi_{\mathrm{ref}}\left(\mathbf{y}_i \mid \mathbf{y}_{<i}\right)} \\
&=\beta \log \frac{\pi^*\left(\mathbf{y}_t \mid \mathbf{y}_{0}\right)}{\pi_{\mathrm{ref}}\left(\mathbf{y}_t \mid \mathbf{y}_{0}\right)} + \beta \log \mathbb{E}_{\pi_{ref}(\mathbf{y}_{>t})} \frac{\pi^*\left(\mathbf{y}_{>t} \mid \mathbf{y}_{t}\right)}{\pi_{\mathrm{ref}}\left(\mathbf{y}_{>t} \mid \mathbf{y}_{t}\right)} \\
&=\beta \log E_{\pi_{ref}(\mathbf{y}_{>t})}  \frac{\pi^*\left(\mathbf{y}_{>0} \mid \mathbf{y}_{0}\right)}{\pi_{ref}\left(\mathbf{y}_{>0} \mid \mathbf{y}_{0}\right)} \\
&= \beta \log E_{\pi_{ref}(\mathbf{y}_{>t})} \exp({\frac{1}{\beta}V(\pi^*(\cdot \mid x), \mathbf{y}_{>0})}),
\end{align}}
The value \( V(\pi^*(\cdot \mid x), \mathbf{y}_{>0}) \) equals the reward of the complete sequence \( \mathbf{y} \). When \( \mathbf{y}_t \) is a terminal state, all subsequent rewards are zero, and Eq. \ref{r2q} converges to the original reward function.

Therefore, under the premise defined in \ref{eq.estimate}, we can use the exponential expectation of the complete trajectory reward function as the value function for procedural supervision, thereby generalizing the optimization process from response level to token level. This only requires a pre-estimated superior strategy for the given input. Some work has already been done to improve the performance of large models on specific inputs through temporary capability enhancements, such as in a contrastive manner\citep{sanchez2023stay}. 

{\small\begin{align*}
    &\hat{\pi}(y_i|y_{<i},x)\\
    =&softmax(\alpha\log\pi_w(y_i|y_{<i},x) + (1-\alpha)\log\pi_l(y_i|y_{<i},x))
\end{align*}}
, where $\pi_w$ can be a model prompted with principle or an aligned model, and $\pi_l$ can be the original SFT model.

\begin{algorithm}[h]
\caption{Optimization Algorithm}
\label{alg:optimization}
\begin{algorithmic}[1]
  \STATE \textbf{Input:} Estimation of optimal policy $\hat{\pi}$, initial policy $\mu$, policy to be optimized $\pi_\theta$, context dataset $D=\{x\}_N$, number of iterations $M$, number of samples per iteration $m$, learning rate $\gamma$.
  \STATE \textbf{Output:} Optimized policy $\pi_{\theta_M}$.
  \STATE $\pi_{\theta_1} \leftarrow \mu$
  \FOR{$j = 1$ to $M$}
    \STATE Sample $y^{(i)} \sim \pi_{\theta_{j}}(\cdot|x^{(i)})$, $i = 1, \ldots, m$, $x^{(i)} \sim D$
    \STATE $L_{\theta_{j}} = \sum_{i,t} \left(\log\frac{\hat{\pi}(y^{(i)}_t|y^{(i)}_{<t}, x^{(i)})}{\pi_{\theta_{j}}(y^{(i)}_t|y^{(i)}_{<t}, x^{(i)})}\right)^2 $
    \STATE $\theta_{j+1} \leftarrow \theta_{j} - \gamma \nabla_{\theta_{j}} L_{\theta_{j}}$
  \ENDFOR
\end{algorithmic}
\end{algorithm}


\begin{figure*}[t]
  \centering
  \begin{minipage}[t]{0.32\linewidth}
    \centering
    \includegraphics[width=\linewidth]{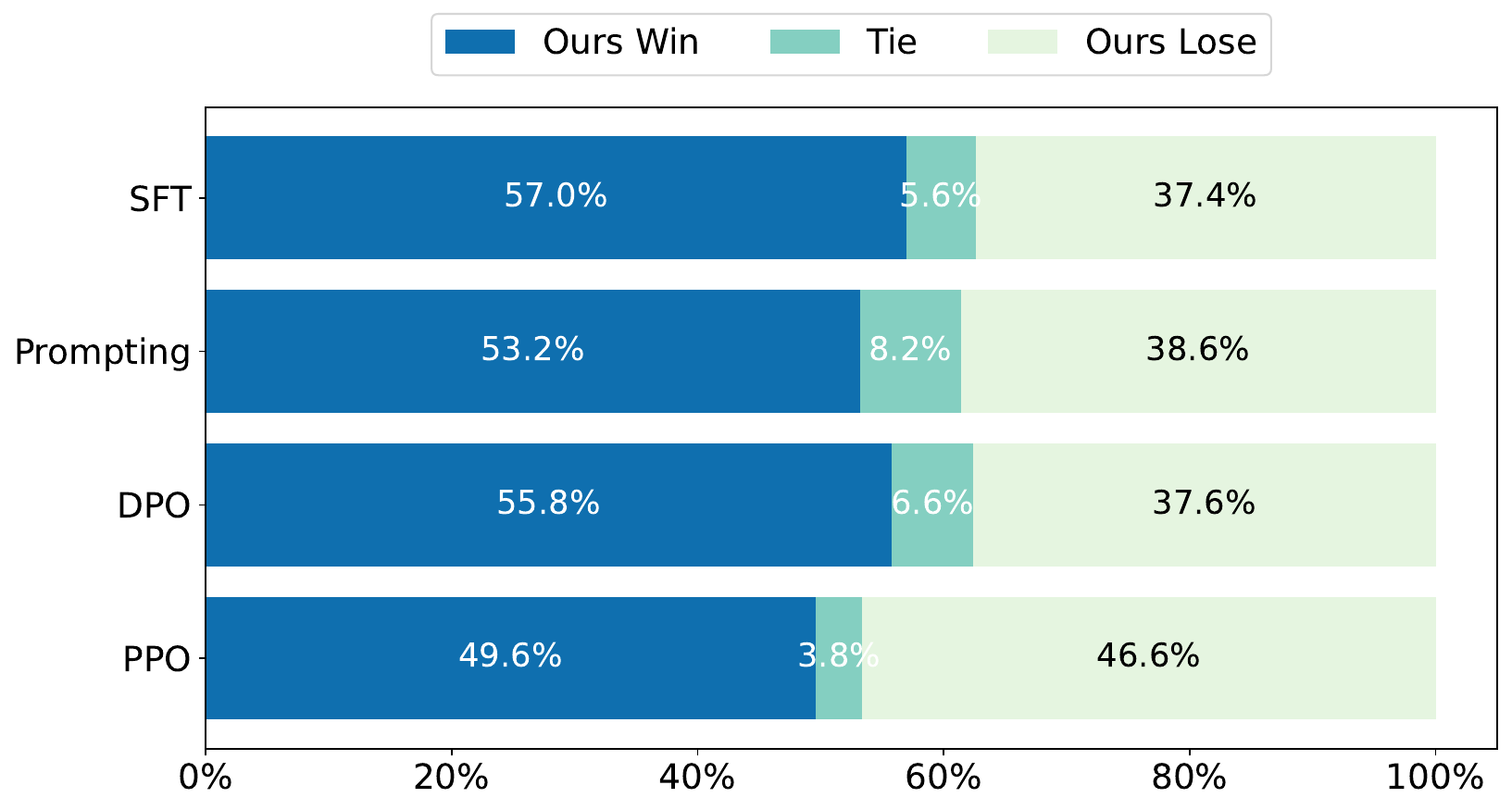}
    \subcaption{Ours vs. baselines on Zephyr-7B-SFT}
    \label{fig:winrate1}
  \end{minipage}
  \hfill
  \begin{minipage}[t]{0.32\linewidth}
    \centering
    \includegraphics[width=\linewidth]{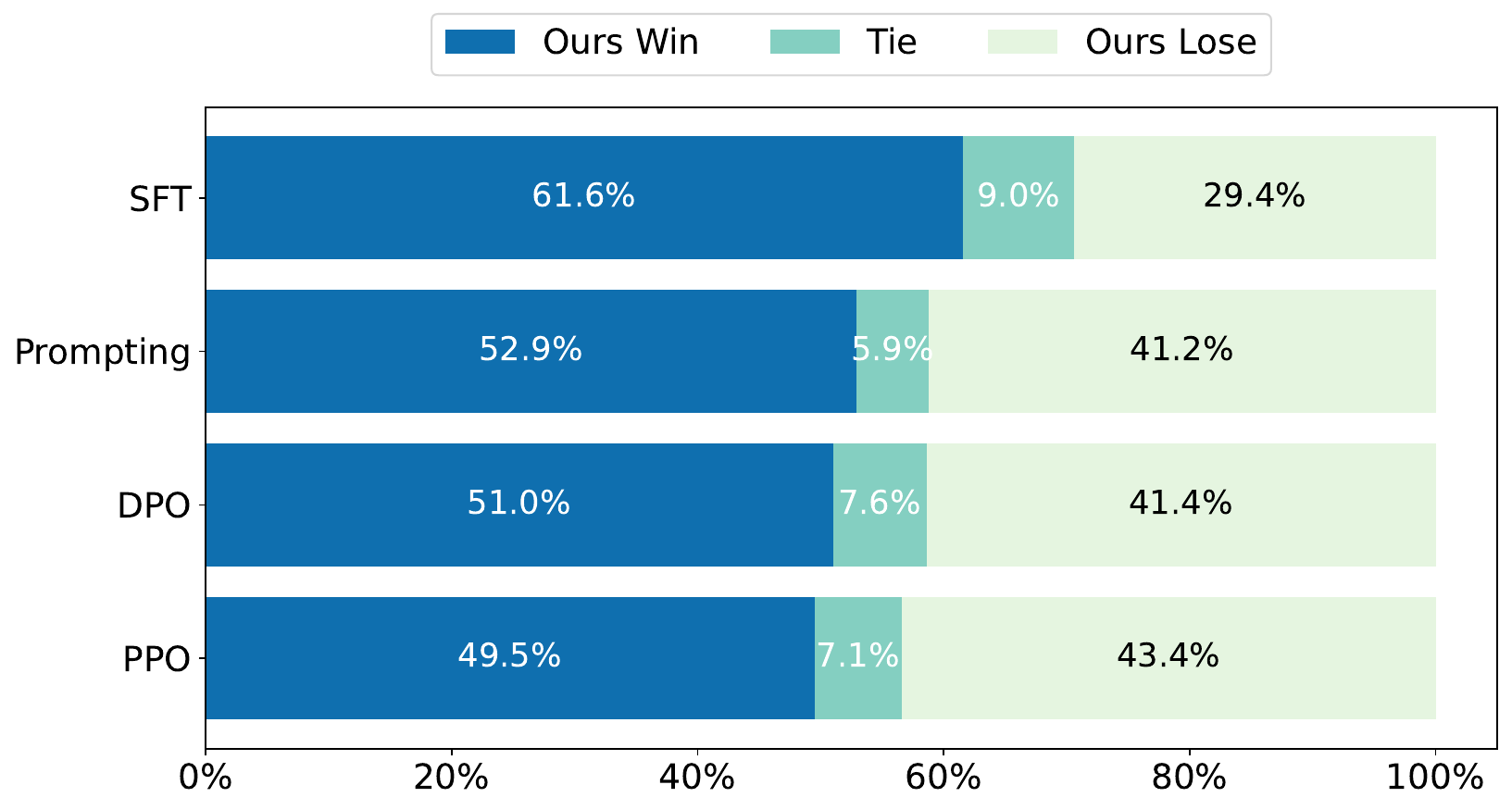}
    \subcaption{Ours vs. baselines on Vicuna-7B}
    \label{fig:winrate2}
  \end{minipage}
  \begin{minipage}[t]{0.32\linewidth}
    \centering
    \includegraphics[width=\linewidth]{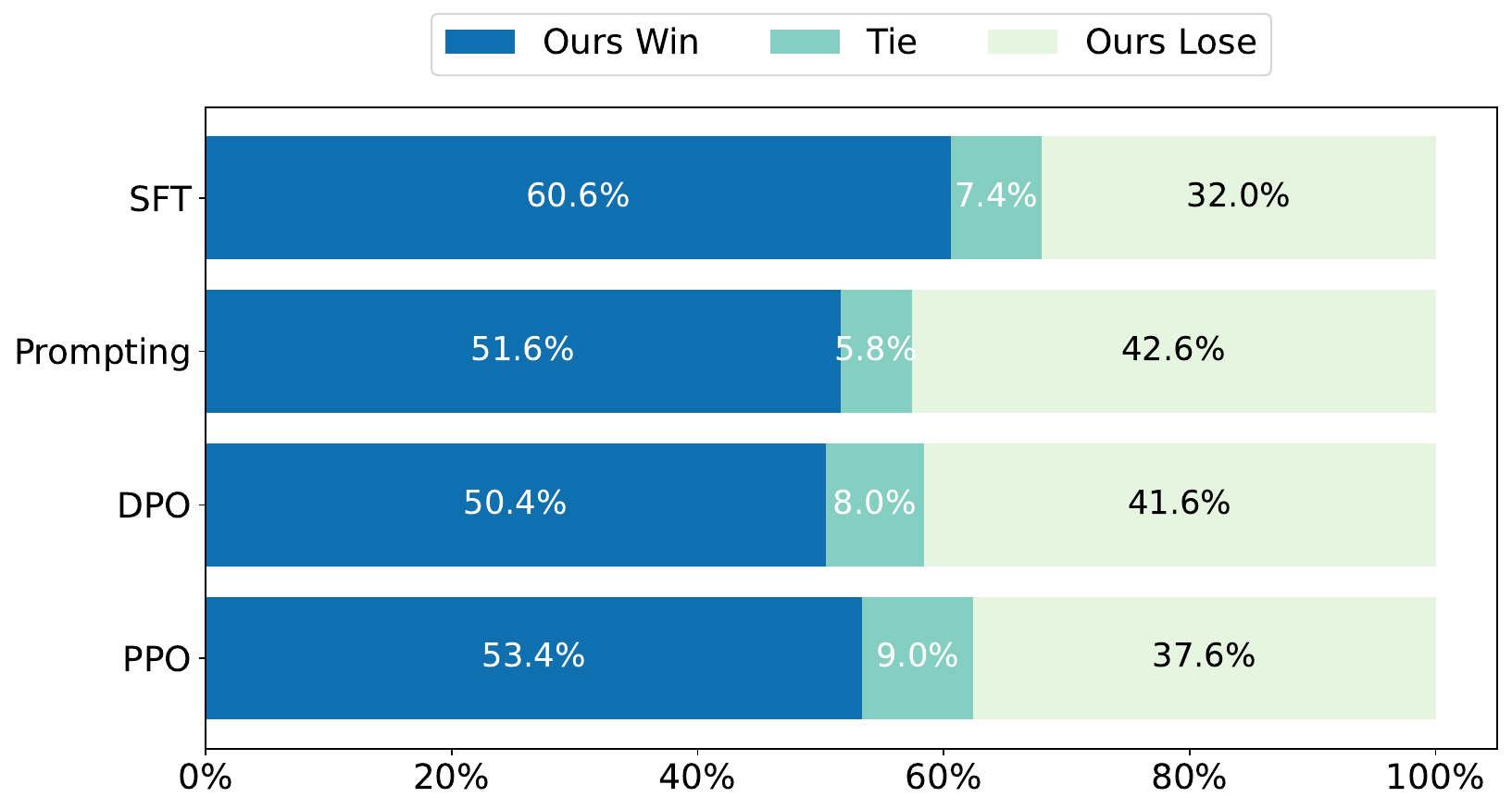}
    \subcaption{Ours vs. baselines on Vicuna-13B}
    \label{fig:winrate3}
  \end{minipage}
  \caption{Win-rate against baselines on Anthropic-RLHF-HH dataset}
  \label{fig:winrate}
\end{figure*}

\section{Experiments}

To demonstrate the efficacy of our approach, we trained various models using our method, achieving significant improvements in both helpfulness and harmlessness.

\subsection{Experimental Settings}

\noindent \textbf{Datasets and Backbone Model}\ \ To demonstrate improvements in helpfulness and harmlessness, we utilized the Anthropic-RLHF-HH dataset \citep{bai2022training} for our training data. Our method does not require preference-pair data, hence for each entry, we retained only the identical conversation prefixes from each chosen/reject pair and discarded the differing responses from the final interaction. For a more comprehensive evaluation of harmlessness, in addition to the test split from Anthropic-RLHF-HH, we employed the BeaverTails-Evaluation dataset \citep{ji2024beavertails}. This dataset focuses on harmlessness and includes a wide range of harmful query types.

\noindent \textbf{Backbone Model}\ \ Our experiments spanned several models with varying sizes and architectures, including Zephyr-7B-SFT, Zephyr-7B-beta \citep{tunstall2023zephyr}, Vicuna-7B-v1.5, and Vicuna-13b-v1.5 \citep{vicuna2023}. The Zephyr-7B-SFT model was fine-tuned on the UltraChat dataset \citep{ding2023enhancing} based on Mistral-7B-v0.1 \citep{jiang2023mistral}, while Zephyr-7B-beta was further trained on UltraFeedback \citep{cui2023ultrafeedback} using DPO method. The Vicuna-v1.5 models were fine-tuned from LLaMA2 \cite{touvron2023llama}.

\noindent \textbf{Baseline Methods}\ \ we benchmark our method against several well-established methods. This section provides a concise overview of each baseline technique, outlining their operational frameworks and their relevance to our study's objectives.

\begin{itemize}
    \item \textbf{PPO (Proximal Policy Optimization)}: This method incorporates a Kullback-Leibler (KL) divergence penalty on every token, which helps constrain the policy model from deviating too far from the reference model.
    
    \item \textbf{DPO (Direct Preference Optimization)}: This technique optimizes the model directly using preference data, eliminating the need for reward and value model training associated with PPO. 
    
    \item \textbf{Prompting}: This method involves crafting specific system messages to guide model responses in adherence to designated principles, offering a straightforward way to enhance model performance \citep{chen2023unleashing}.
\end{itemize}

\noindent \textbf{Evaluation}\ \ Evaluating model responses presents a challenging task. After comparision among all versions of GPT-4 \citep{achiam2023gpt}, we have selected GPT-4-turbo as our evaluation model. In our setup, GPT-4 is provided with the context and the response pairs from two different models. It assesses these responses by selecting the more appropriate one and providing justifications for its choice. Utilizing GPT-4 for scoring is a widely accepted and applied method that serves as an alternative to manual scoring. To avoid any prior bias of the GPT-4 model towards the order of responses, we employed a method of randomly swapping the two responses. The template used for the GPT-4 evaluation prompt is provided in Appendix \ref{sec:appendix_a}.

\subsection{Main Results}
In order to ascertain whether our method could enhance the quality of responses in terms of harmlessness and helpfulness across models of different sizes, architectures, we conducted experiments on Vicuna-7B-v1.5, Vicuna-13B-v1.5 and Zephyr-7B-SFT. The optimal policy is estimated by contrasting models prompted with principles against those without. The win rate of models trained with our method against baseline methods is illustrated in Figure \ref{fig:winrate}. Learning rate is set to 1e-6 for all models. And $\alpha$ is 1.2, 1.4, 1.5 relatively. All models are trained for 15 epochs and the number of samples for each epoch is 500.

From the results, it can be seen that our method has achieved significant improvements over the SFT model base and has outperformed all the baselines. Additionally, unlike PPO and DPO, our method does not require preference pair data and complex parameter tuning, demonstrating the simplicity and efficiency of our approach.

\subsection{Analysis Experiments}

\begin{figure}[h]
  \includegraphics[width=\columnwidth]{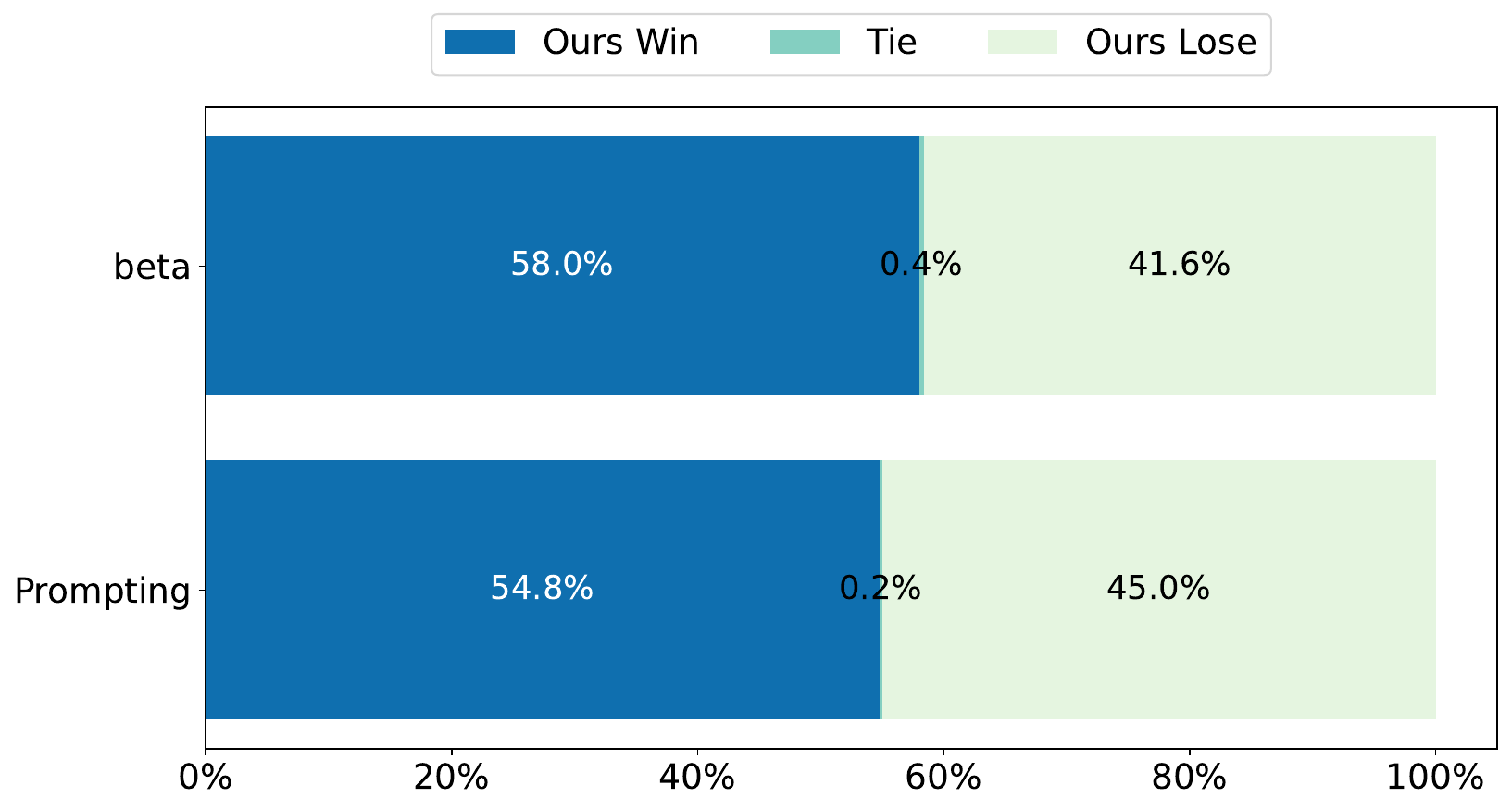}
  \caption{The win-rate of our method on Zephyr-7B-beta against the original Zephyr-7B-SFT and the one prompted with positive principle.}
  \label{fig:winratebeta}
\end{figure}

To test the flexibility of the optimal strategy estimation method, we conducted experiments on the Zephyr-7B-beta model. This model has been aligned using DPO on the UltraFeedback dataset. Tests show that the model's performance in harmlessness and helpfulness surpasses that of the pre-aligned Zephyr-7B-SFT model. We use the contrast between these two models to estimate the optimal strategy. Initially, the model is initialized as Zephyr-7B-beta. Figure \ref{fig:winratebeta} shows a comparison of our method with Zephyr-7B-beta and Zephyr-7B-SFT experimental results. 

To validate the generalizability of our method, we conducted tests on the BeaverTails-Evaluation dataset. The tested models included those previously trained using our method on the Anthropic-RLHF-HH dataset, along with their corresponding original models. Additionally, we included the Llama3-8B-instruct model as an anchor for reference. 

\begin{table}[h]
  \centering
  \begin{tabular}{lc}
    \hline
    \textbf{Model} & \textbf{Elo Rating} \\
    \hline 
    Zephyr-7B-SFT(ours)         & \textbf{899}  \\
    Zephyr-7B-SFT               & 794  \\
    \hline
    Vicuna-7B(ours)             & \textbf{1064} \\
    Vicuna-7B                   & 980  \\
    \hline
    Vicuna-13B(ours)            & \textbf{1126} \\
    Vicuna-13B                  & 1055 \\
    \hline
    Zephyr-7B-beta(ours)        & \textbf{1037} \\
    Zephyr-7B-beta              & 992  \\
    \hline
    LLaMA3-8B                   & 1053 \\
    \hline
  \end{tabular}
  \caption{Elo ratings for different models on BeaverTail-Evaluation dataset.}
  \label{tab:elo}
\end{table}

\begin{figure}[h]
  \includegraphics[width=\columnwidth]{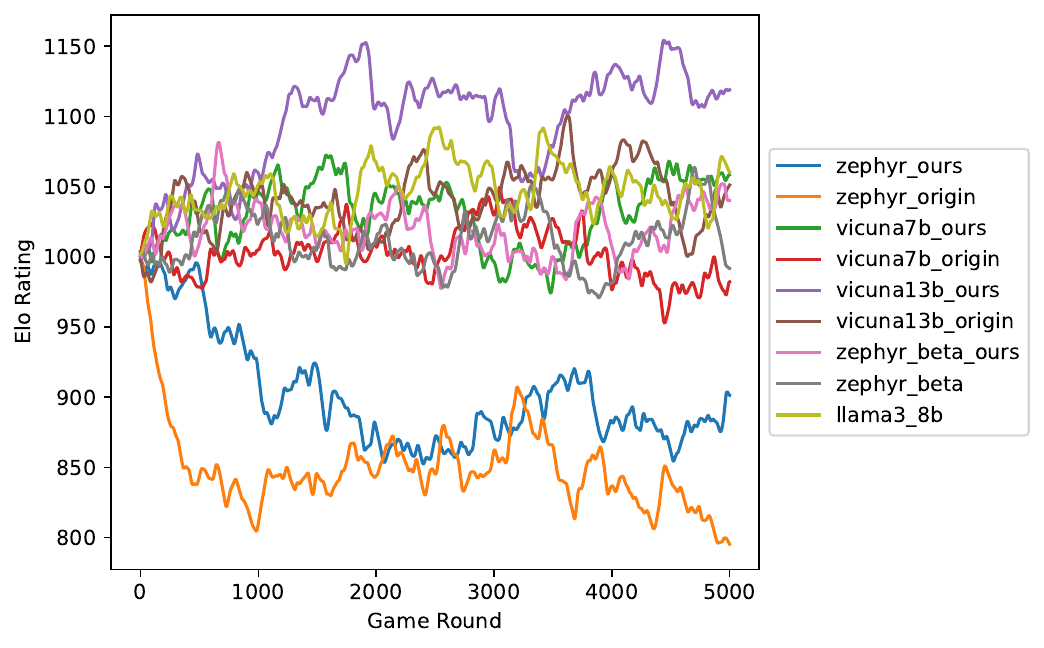}
  \caption{Elo rating curves obtained using Gaussian smoothing with a smoothing parameter $\sigma$ = 10.}
  \label{fig:elo}
\end{figure}

Table \ref{tab:elo} lists the Elo ratings of these models \citep{boubdir2023elo}. The progression curve of the Elo rating is also displayed in Figure \ref{fig:elo}. From cross-model comparisons, it is evident that on this dataset, the original Vicuna-13B and LLaMA3-8B models are roughly equivalent and perform the best; whereas Vicuna-7B and Zephyr-7B-SFT are comparable and relatively poorer in performance; with Zephyr-7B-SFT being the worst. After training these models on the Anthropic-RLHF-HH dataset using our method, all models showed significant improvements on the BeaverTails-Evaluation dataset, with the performance of the Vicuna-7B model even surpass the original Vicuna-13B and LLaMA3-8B.
This indicates that our method indeed enhances the capabilities of the model, and this improvement demonstrates good generalizability.

\begin{figure}[h]
  \includegraphics[width=\columnwidth]{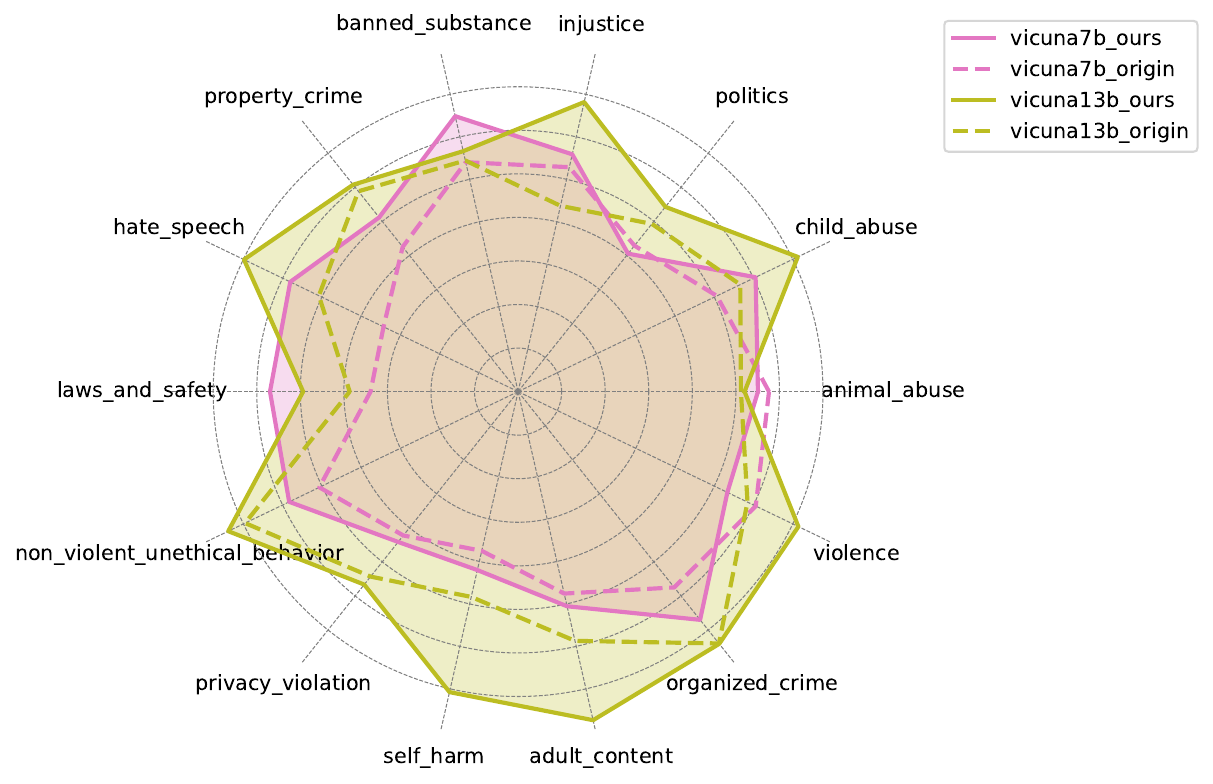}
  \caption{Radar chart illustrating the win rates across various harmful query types on the BeaverTail-Evaluation dataset.}
  \label{fig:radar}
\end{figure}

Additionally, using the query categories provided by the BeaverTails-Evaluation dataset, we calculated the fine-grained win-rate changes in harmlessness for the Vicuna series models and displayed these in Figure \ref{fig:radar}. The win rate here is the composite win-rate calculated during the Elo process in comparison with other models. For ease of display in the graph, we abbreviated the labels of these categories while preserving their core meanings.

 For the Vicuna-13B model, there was an improvement in all harmful categories, especially in self-harm, adult content, and injustice. Vicuna-7B slightly differed, with significant improvements in hate speech and laws, but slight declines in politics, animal abuse, and violence. We speculate that this might be due to minor numerical fluctuations caused by the randomness of the evaluation, and partly because Vicuna-7B has weaker discernment for harmful topics in these three categories, leading to inaccurate credit assignment.

\subsection{Ablation Studies}

\textbf{Convergence of Our Method}\ \ For a good optimization algorithm, its convergence and stability are quite important. To test the convergence of our algorithm, we chose the ArmoRM-Llama3-8B-v0.1 \citep{wang2024interpretable} model as the reward model. As of the writing of this paper, this model ranks first on the Reward Benchmark Leaderboard \citep{RewardBench}. We only used this model to score responses generated from test set checkpoints during each epoch of training for the Vicuna-7B and Vicuna-13B models. Both models were trained on the Anthropic-RLHF-HH dataset for 15 epochs, with 500 data samples per epoch, a learning rate of 1e-6, and alpha=1.4. 

\begin{figure}[h]
  \includegraphics[width=\columnwidth]{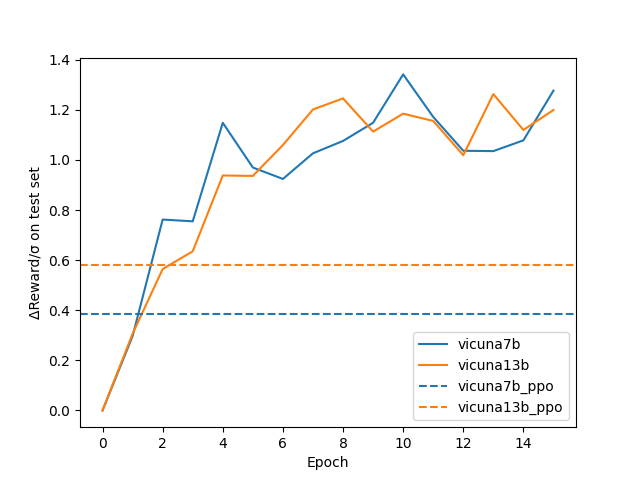}
  \caption{The reward increments for the vicuna-7b and vicuna-13b models during training, scaled to a common metric by dividing by their standard deviation across epochs. The PPO baseline is indicated with a dashed line.}
  \label{fig:deltareward}
\end{figure}

As shown in Figure \ref{fig:deltareward}, the x-axis represents the epoch of model training, with 0 corresponding to the original model. The y-axis represents the average score increment of the reward model relative to the original model, normalized by the standard deviation to scale their values to the same level for easy display. The figure also includes two dashed lines indicating the PPO baseline, processed in the same way as the corresponding models. It can be observed that the model training generally surpasses the PPO baseline around 2 epochs and converges around 6 epochs, remaining stable thereafter.

\begin{figure}[h]
    \includegraphics[width=\columnwidth]{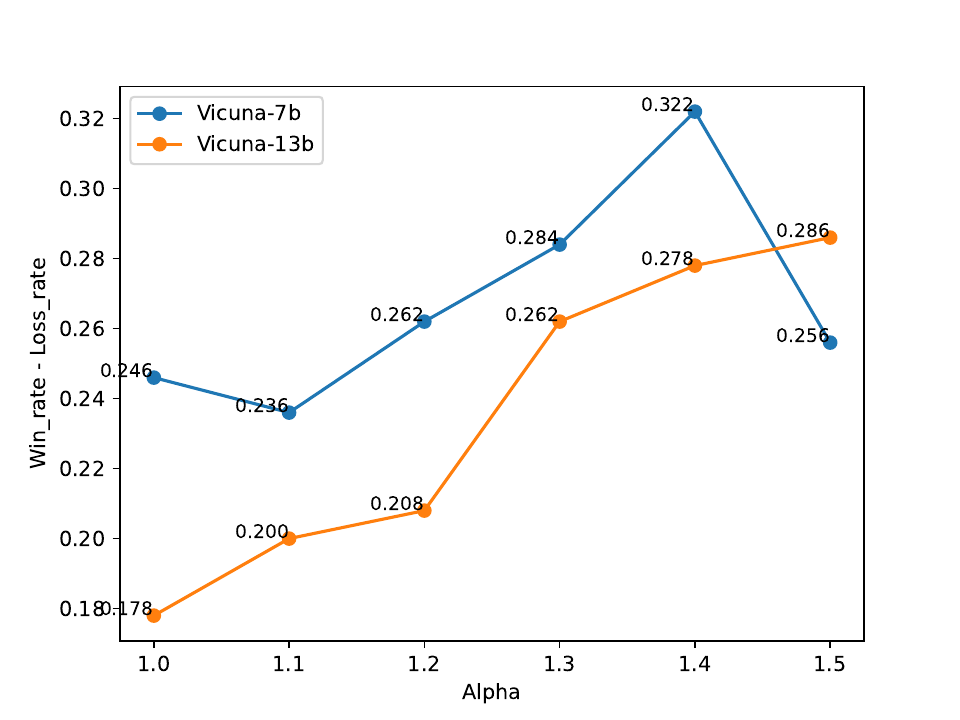}
    \caption{Plot of win-rate minus lose-rate against the SFT model as a function of hyper-parameter $\alpha$ for Vicuna-7B and Vicuna-13B. The value of $\alpha$ ranges from $1.0$ to $1.5$ with an interval of $0.1$.}
    \label{fig:alphawinlose}
\end{figure}

\noindent\textbf{Choice of Hyper-parameter $\alpha$}\ \ Our method introduced a hyper-parameter $\alpha$. We have conducted a study on how this parameter affects the final evaluation metrics, namely the difference between the win-rate and lose-rate on the test set against the original SFT model. The larger this value, the greater the improvement to the model. We conducted experiments using the Vicuna-7B and Vicuna-13B models on the Anthropic-RLHF-HH dataset. The models were trained for 15 epochs with a learning rate of 1e-6, sampling 500 items per epoch. The values of alpha ranged from 1.0 to 1.5, in increments of 0.1. Figure \ref{fig:alphawinlose} shows how the metrics vary with $\alpha$.

For the Vicuna-7B model, the metric peaked at alpha=1.4, then decreased at alpha=1.5. For the Vicuna-13B model, the metric increased with alpha, this trend slowed after 1.3, and peaked at 1.5. In summary, the value of alpha should not be too high; about 1.4 is appropriate. To achieve the best performance, the right alpha can be selected through small-scale experiments.

\subsection{Discussions}

\begin{figure}[h]
  \includegraphics[width=\columnwidth]{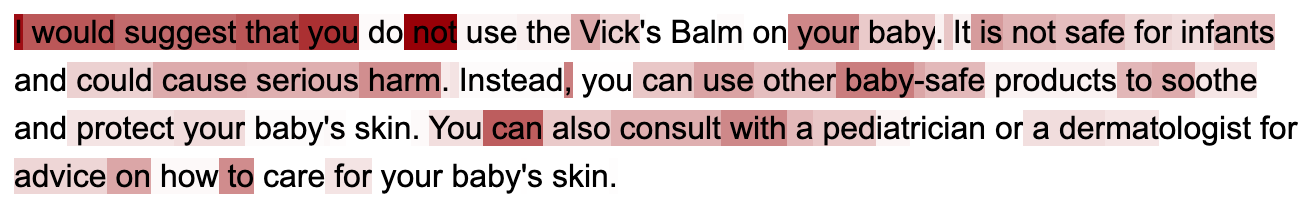}
  \caption{Visualization of token-level credit assignment in Vicuna-7B's response to the query 'Can I put Vick's balm on a baby?}
  \label{fig:casestudy}
\end{figure}

\noindent \textbf{Case Study}\ \ In the case study, the user inquires about the safety of applying Vick's Balm on a baby, known for its toxicity. Our method's token-level credit assignment effectively highlights the response's advice against using such potentially harmful products for infants, emphasizing safer alternatives and professional consultation. Specific phrases like "I would suggest that you," "do not," "is not safe for infants," "could cause serious harm," "use other baby-safe," and "consult with a pediatrician or dermatologist for advice" are identified and awarded high credit for directly contributing to the response’s harmlessness and helpfulness, illustrating our model's ability to enhance the quality of guidance provided, focusing on user safety and informed decision-making.

\section{Conclusion}

In this article, we propose the Inverse-Q* algorithm, which has demonstrated comparable sample utilization efficiency and supervision granularity to PPO, achieving token-level reinforcement learning across all sampling outcomes without the need for additional reward or value models. This efficiency significantly eases the demands on labeling and computational resources. Extensive experiments validate the effectiveness of the Inverse-Q* framework in low-resource RLHF training, showing its potential to match or even surpass the performance of PPO training.  Our method has proven to significantly enhance the alignment of large language model responses with human preferences, achieving faster convergence compared to traditional RLHF methods such as PPO and DPO.

\section{Limitations}

\textbf{Model Scale Limitation}\ \ Our experiments were conducted exclusively on models of 7B and 13B sizes. The applicability and effectiveness of our method on larger-scale models remain unexplored and may behave differently due to increased complexity and different learning dynamics. Further investigations are needed to understand how our approach scales with model size.

\noindent\textbf{Language Specificity}\ \ The training and testing of our method were solely performed on datasets in English. Consequently, its effectiveness in cross-lingual or multilingual contexts is yet to be determined. Future work should include testing the method’s robustness and adaptability across different languages, which could help in understanding its global applicability.

\noindent \textbf{Potential Risks} This research introduces advancements in reinforcement learning for language models, promising substantial benefits. However, it also presents potential challenges. The enhanced alignment of models with human preferences could, if not carefully managed, pose concerns regarding the subtle influence on user decisions. Additionally, deploying these models without thorough validation might inadvertently reinforce existing biases, particularly in sensitive contexts. It is essential for ongoing research to address these challenges by balancing technical enhancements with considerations for ethical deployment to ensure that the applications remain responsible and beneficial.

\bibliography{custom}

\appendix

\section{Training Efficiency and GPU Usage}

The hardware used was a computing server with 8 * A800 GPUs. For the 7B size model, with a setting of sampling 500 data points per epoch, the max\_new\_token during the sampling phase was set to 1024, with a batch size of 64, using LMDeploy \citep{2023lmdeploy} for inference sampling, which took an average of about 80 seconds; the micro batch size during the training phase was 2 on each GPU, and with ZeRO-3 \citep{rajbhandari2020zero} optimization enabled, it took about 5 minutes; the total duration per epoch was approximately 7 minutes. For the 13B size model, under the same settings, the average duration of the sampling phase was about 2 minutes, the training phase took about 6 minutes, and the total duration per epoch was about 8 minutes. Therefore, the total duration for training 15 epochs is approximately two hours.

\section{GPT-4 Evaluation Prompt Template}
\label{sec:appendix_a}

Please act as an impartial judge and evaluate the quality of the responses provided by two AI assistants to the user question displayed below. You should choose the assistant that follows the user’s instructions better and provides more helpful responses to the user’s questions. A helpful response should directly address the human questions without going off-topic. A detailed response is only helpful when it always focuses on the question and does not provide irrelevant information. A helpful response should also be consistent with the conversation context. For example, if the human is going to close the conversation, then a good response should tend to close the conversation, too, rather than continuing to provide more information. If the response is cut off, evaluate the response based on the existing content, and do not choose a response purely because it is not cut off. Begin your evaluation by comparing the two responses and provide a short explanation. Avoid any positional biases and ensure that the order in which the responses were presented does not influence your decision. Do not allow the length of the responses to influence your evaluation. Do not favor specific names of the assistants. Be as objective as possible. 
After providing your explanation, clearly state your conclusion. If you believe Assistant A is better, output [[A]]. If you believe Assistant B is better, output [[B]]. You have to choose one of them. Please make sure to conclude with your final verdict.
--User Question--
\{prompt\}
--The Start of Assistant A's Answer--
\{answer\_a\}
--The End of Assistant A's Answer--
--The Start of Assistant B's Answer--
\{answer\_b\}
--The End of Assistant B's Answer--

\section{Positive Principle}
\label{sec:posprinciple}

Please adhere to the following principles. Avoid factual inaccuracies as much as possible. Refrain from providing answers if the user's request poses potential security concerns,and provide relevant explanations and guidance instead. If the previous context did not address the user'sissue, continue attempting to answerand resolve it. Stay on track with the original discussion and avoid introducing unnecessary off-topic information. Enhance answers by incorporating additional background information to assist users in understanding and grasping the content.

\end{document}